\documentclass[10.5pt,compsoc]{tst}
\usepackage{graphicx}
\usepackage{footmisc}
\usepackage{subfigure}
\usepackage{url}
\usepackage{multirow}
\usepackage[noadjust]{cite}
\usepackage{amsmath,amsthm}
\usepackage{amssymb,amsfonts}
\usepackage{booktabs}
\usepackage{color}
\usepackage{ccaption}
\usepackage{booktabs}
\usepackage{float}
\usepackage{fancyhdr}
\usepackage{caption}
\usepackage{xcolor,stfloats}
\usepackage{comment}
\graphicspath{{figures/}}
\usepackage{cuted}  
\usepackage{captionhack}
\usepackage{epstopdf}

\usepackage{algorithmic}
\usepackage{algorithm}
\usepackage{array}

\usepackage{makecell}
\usepackage{multirow}

\usepackage{orcidlink}

\usepackage{colortbl}

\headevenname{\zihao{-5}{\textbf{\emph{Tsinghua Science and Technology, February}} 2018, 23(1): 000--000}}%
\headoddname{{\sf First author et al.:}\quad {\textbf{\emph{Click and Type Your Title Here ...}}}}%

\newtheoremstyle{mystyle}{0pt}{0pt}{\normalfont}{1em}{\bf}{}{1em}{}
\theoremstyle{mystyle}
\renewcommand\figurename{Fig.~}

\newcommand{\nop}[1]{}

\makeatletter
\renewcommand{\@biblabel}[1]{[#1]\hfill}
\makeatother
\begin{document}

\thispagestyle{empty}



\hyphenpenalty=50000

\makeatletter
\newcommand\mysmall{\@setfontsize\mysmall{7}{9.5}}

\newenvironment{tablehere}
  {\def\@captype{table}}
  {}
\newenvironment{figurehere}
  {\def\@captype{figure}}
  {}

\thispagestyle{plain}%
\thispagestyle{empty}%

\let\temp\footnote
\renewcommand \footnote[1]{\temp{\zihao{-5}#1}}
{}
\vspace*{-40pt}
\noindent{\zihao{5-}\textbf{\scalebox{0.885}[1.0]{\makebox[5.9cm][s]
{TSINGHUA\, SCIENCE\, AND\, TECHNOLOGY}}}}

\vskip .2mm
{\zihao{5-}
\textbf{
\hspace{-5mm}
\scalebox{1}[1.0]{\makebox[5.6cm][s]{%
I\hfill S\hfill S\hfill N\hfill{\color{white}%
l\hfill l\hfill}1\hfill0\hfill0\hfill7\hfill-\hfill0\hfill2\hfill1\hfill4
\hfill \color{white}{\quad 0\hfill ?\hfill /\hfill ?\hfill ?\quad p\hfill p\hfill  ?\hfill ?\hfill ?\hfill --\hfill ?\hfill ?\hfill ?}\hfill}}}}

\vskip .2mm
{\zihao{5-}
\textbf{
\hspace{-5mm}
\scalebox{1}[1.0]{\makebox[5.6cm][s]{%
DOI:~\hfill~\hfill1\hfill0\hfill.\hfill2\hfill6\hfill5\hfill9\hfill9\hfill/\hfill T\hfill S\hfill T\hfill.\hfill2\hfill0\hfill1\hfill8\hfill.\hfill9\hfill0\hfill1\hfill0\hfill0\hfill0\hfill0}}}}

\vskip .2mm\noindent
{\zihao{5-}\textbf{\scalebox{1}[1.0]{\makebox[5.6cm][s]{%
\color{white}{V\hfill o\hfill l\hfill u\hfill m\hfill%
e\hspace{0.356em}1,\hspace{0.356em}N\hfill u\hfill%
m\hfill b\hfill e\hfill r\hspace{0.356em}1,\hspace{0.356em}%
S\hfill e\hfill p\hfill t\hfill e\hfill%
m\hfill b\hfill e\hfil lr\hspace{0.356em}2\hfill0\hfill1\hfill8}}}}}\\

\begin{strip}
{\center
{\zihao{3}\textbf{
Attention-aggregated Attack for Boosting the Transferability of \\Facial Adversarial Examples}}
\vskip 9mm}

{\center {\sf \zihao{5}
    Jian-Wei Li, Wen-Ze Shao$^*$
}
\vskip 5mm}
%

\centering{
\begin{tabular}{p{160mm}}

{\zihao{-5}
\linespread{1.6667} %
\noindent
\bf{Abstract:} {\sf
Adversarial examples have revealed the vulnerability of deep learning models and raised serious concerns about information security. The transfer-based attack is a hot topic in black-box attacks that are practical to real-world scenarios where the training datasets, parameters, and structure of the target model are unknown to the attacker. However, few methods consider the particularity of class-specific deep models for fine-grained vision tasks, such as face recognition (FR), giving rise to unsatisfactory attacking performance. In this work, we first investigate what in a face exactly contributes to the embedding learning of FR models and find that both decisive and auxiliary facial features are specific to each FR model, which is quite different from the biological mechanism of human visual system. Accordingly we then propose a novel attack method named Attention-aggregated Attack (AAA) to enhance the transferability of adversarial examples against FR, which is inspired by the attention divergence and aims to destroy the facial features that are critical for the decision-making of other FR models by imitating their attentions on the clean face images. Extensive experiments conducted on various FR models validate the superiority and robust effectiveness of the proposed method over existing methods.}
\vskip 4mm
\noindent
{\bf Key words:} {\sf deep neural networks; face recognition; adversarial transferability; intermediate-level attacks}}

\end{tabular}
}
\vskip 6mm

\vskip -3mm
\zihao{6}\end{strip}

\thispagestyle{plain}%
\thispagestyle{empty}%
\makeatother
\pagestyle{tstheadings}

\begin{figure}[b]
\vskip -6mm
\begin{tabular}{p{44mm}}
\toprule\\
\end{tabular}
\vskip -4.5mm
\noindent
\setlength{\tabcolsep}{1pt}
\begin{tabular}{p{1.5mm}p{79.5mm}}

$\bullet$& Jian-Wei Li and Wen-Ze Shao are with Jiangsu Key Laboratory of Intelligent Information Processing and Communication Technology, School of Communications and Information Engineering, Nanjing University of Posts and Telecommunications, Nanjing 210003, P.R. China (e-mail:{1022010429, shaowenze}@njupt.edu.cn).\\

$\sf{*}$&
To whom correspondence should be addressed. \\
          &          Manuscript received: year-month-day;
          accepted: year-month-day

\end{tabular}
\end{figure}\zihao{5}

\vbox{}
\vskip 1mm
\noindent
\section{Introduction}
Facial recognition (FR) technology has made tremendous progress to date, enabling numerous applications in various complex systems such as facial payment, access control, and video surveillance. Despite the prosperity achieved in facial recognition technology, the state-of-the-art FR models are suffering from serious information security concerns that they are vulnerable to adversarial examples crafted by adding imperceptible perturbation to benign images. It is of great necessity to study the vulnerability of FR models.

In white-box setting, the training datasets, structures, and parameters of the target model are exposed to attackers for generating adversarial examples against the target model with extremely high attack success rates (ASRs). However, in more practical black-box scenarios, attackers only have access to the final decision of the target model, which is challenging for realizing satisfactory attacking performance. Query-based black-box attacks, largely classified into two categories: interacting with the target model \cite{quert-tar1,quert-tar2,quert-tar3} or training a substitute model \cite{DMA,DML}, require a large number of queries to generate reliable adversarial examples, suffering from limited query efficiency. Transfer-based attacks, another booming direction of black-box attacks and characterized by great attacking efficiency, craft adversarial examples on a surrogate model. Thus, the commonality in decision-making shared between the surrogate model and target model is important for achieving strong transferable attacks. 

Many efforts have been made to promote transfer-based attacks, such as data augmentation \cite{DIM, NI-SIM}, feature-level destruction \cite{FIA, NAA}, and so on. Few works are specifically objected to the FR task. One example is the LGC \cite{LGC}, which is targeted to attack FR models and takes the particularity of FR models into consideration by randomly occluding the prior facial landmarks of a face image in each attacking iteration, intending to transfer the attention of a source model to other unnoticed facial features that may be critical for the decision-making of target FR models. However, occluding facial landmarks in LGC does not align well with its objective of transferring attention, failing to destroy the vast majority of facial features, and thus contributing to unsatisfactory adversarial transferability. 

Many studies have delved into face recognition in psychology and neuroscience. For example, \cite{face-photo} found that some facial features play decisive roles while other features are auxiliary in face verification. However, it is uncertain what exactly in a face contributes to the embedding learning of FR models, and we wonder if all facial features participate in the decision-making, or if FR models prefer to recognize some specific features or follow a similar biological mechanism of human visual system. In this study, it is observed that different FR models have their own characteristics. That is, the decisive and auxiliary facial features are just specific to each FR model, from which we can infer that the adversarial attack on a source model tends to destroy its decisive and auxiliary features, yet inevitably prone to overfitting the adversarial examples to that model, and thus degrading the adversarial transferability. To address this problem, a novel method named Attention-aggregated attack (AAA) is proposed, which aims to destroy the facial features that are critical for target FR models by imitating their attentions on the clean face images. The main contributions of this study are as follows:

(1) It is observed that the decisive and auxiliary facial features are specific to FR models, and we accordingly propose the AAA method intuitively inspired by the attention divergence of iterating facial adversarial examples produced by the normal domain-agnostic attack methods, e.g., FGSM-style ones (fast gradient sign method) \cite{DIM,NI-SIM,TIM}.

(2) The AAA method is specifically formulated into a novel feature-level attack guided by visual importance in the form of aggregated attentions, which are naively calculated using off-the-shelf iterating adversarial examples. Extensive experiments conducted on various FR models demonstrate the effectiveness and superiority of our method.

\section{Attention-aggregated Attack}

\subsection{Adversarial Attack against Face Recognition}
Given a deep face model $f_\theta $ parameterized by $\theta$ and a face image pair $\{x^{i},x^{r}\}$, the distance between the high-dimensional embeddings corresponding to the input image face $x^{i}$ and the reference image $x^{r}$ respectively is calculated via $\mathcal{D}_{f_\theta}(x^{i},x^{r})$ where $\mathcal{D}(\cdot)$ is selected as the cosine similarity metric in this study. Ideally, the distance between negative face pairs is larger than positive ones. To explore the vulnerability of deep models, numerous attacking methods such as MIM \cite{MIM} and DIM \cite{DIM}, have been proposed to craft imperceptible adversarial noise $\epsilon$ added to the benign face image. Specifically, taking the dodging attack (positive face pairs) as an example, the optimization objective is given as follows:
\begin{equation}
    \mathrm {arg}\max_{x^{adv}} \mathcal{D}_{f_\theta}(x^{adv},x^{r}),\; s.t.\left \| x^{adv}-x^{i} \right \|_p<\epsilon 
\end{equation}
where $\epsilon$ is regularized by an $\ell_p$-norm, $p$ is set to $\infty$ in this study. 

\textbf{Output-level Attack:}
FGSM \cite{FGSM} is the first attacking method for gaining insight into the vulnerability of deep models. However, its multi-small-step version, i.e., BIM \cite{BIM}, performs worse in transferable attack, which leads the optimization to getting stuck in sub-optimal local minima. Then, momentum was introduced into BIM to alleviate such issue, denoted as MIM \cite{MIM}, and the calculation process of it can be written as:
\begin{equation}
    g_{t+1}=\mu \cdot g_t +\frac{\bigtriangledown _{x^{adv}_t}\mathcal{D}_{f_\theta}(x^{adv}_t,x^{r})}{\left \| \bigtriangledown _{x^{adv}_t}\mathcal{D}_{f_\theta}(x^{adv}_t,x^{r}) \right \|_1 } 
    \label{MI-FGSM}
\end{equation}
\begin{equation}
    x^{adv}_{t+1}=x^{adv}_{t}+\alpha \cdot sign(g_{t+1})
    \label{MI-update}
\end{equation}
where $\mu$ is the momentum and $sign(\cdot)$ represents the symbolic function. To enhance the transferable attack, many variants, such as DIM \cite{DIM}, TIM \cite{TIM}, SIM \cite{NI-SIM}, and beyond, have been subsequently developed. With the deepening insight into adversarial transferability, various attack methods \cite{PGN,STM,qinghua1} are proposed to prevent adversarial examples from being trapped into inferior local optima.
\begin{figure*}[!t]
    \centering
    \includegraphics[width=1.8\columnwidth]{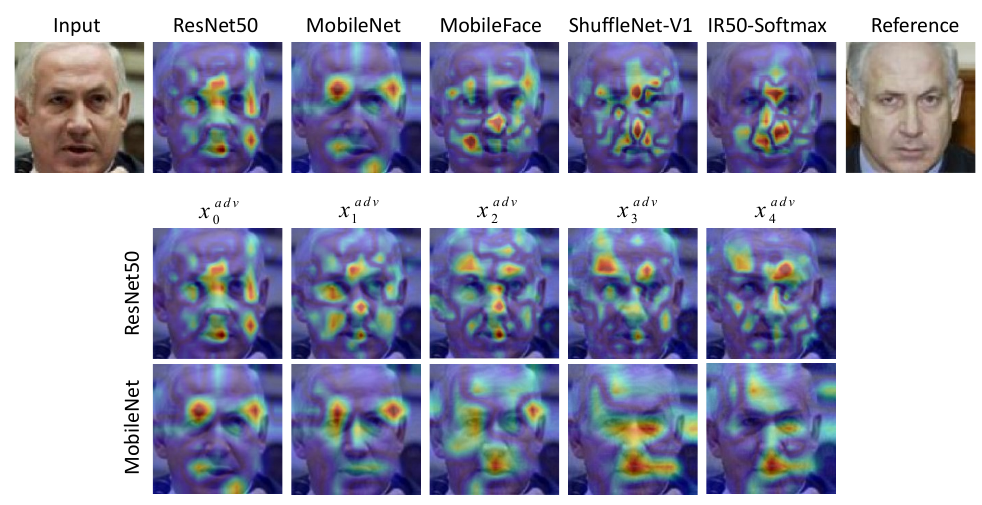}
    \caption{ \textbf{Visualization of the up-sampled gradients (visual importance) of the loss function (cosine similarity) w.r.t. the feature maps in certain mid-layers, which are overlying on the input face image. We select layer2.3 in ResNet50, layer.7 in MobileNet. It can be found that different FRs concentrate on distinct preferable facial features. Gradients of $x^{adv}_k$ displayed in the second and third rows indicate that the attention of a specific model keeps transferring with iteratively added adversarial noise. Considering that the facial features are orthogonal to each other in the pixel space, averaging these diverted visual importance achieves a special kind of ensemble-like effect.}}
    \label{attention}
    \vspace{3mm}
\end{figure*}

\textbf{Feature-level Attack:}
For deep models, it is found that features in a certain mid-layer across different models exhibit higher similarity than shallow and deep layers \cite{mid-layer-similarity}, indicating that attacking feature maps in a certain mid-layer may attain stronger adversarial examples. The more recent advanced feature-level attacks such as FIA \cite{FIA} and NAA \cite{NAA}, are carefully designed to eliminate the noised or semantically non-critical features in the feature maps, while destroying the semantically critical features to boost adversarial transferability. The attacking objective is as follows:
\begin{equation}
    L(x^{adv})=\sum (I\odot h^{adv})
    \label{feature_attack}
\end{equation}
where $\odot$ denotes the element-wise product, $h^{adv}$ represents the mid-layer feature maps extracted from the desired adversarial example, and the attacking direction for optimization is denoted as $I$. Considering the superior performance achieved in feature-level attacks, it is imperative to request and require a specific $I$ adapting the particularity of FR.
\subsection{Attention-aggregated Attack}
\label{AAA}



Numerous studies have addressed various issues about face recognition and discussed the perception of face in psychology and neuroscience. For example, it has been found that although both holistic face and separate facial features are important for face verification, distinguishable facial features dominantly contribute to the recognition \cite{holistic}. Meanwhile, some studies try to investigate which visual features (eyes, nose, mouth, etc.) are more critical in face verification. For example, \cite{face-photo} sorts the visual importance of (or attention on) features according to the psychology of humans (forehead $>$ eyes $>$ mouth $>$ chin $>$ nose). However, which features are critical for FR models are left unknown to date, and we wonder if various FR models follow a similar behavioral pattern in sorting visual importance of these features. To this end, we use the gradients of the cosine similarity w.r.t. feature maps in certain mid-layers calculated on different FR models to qualitatively reflect the visual importance. As shown in Fig. \ref{attention}, different FR models seem to assign distinct visual importance to these facial features, indicating that the decisive and auxiliary features are specific to FR models, rather than following a common biological mechanism of human visual system. It is to say different FR models may recognize a person by different facial features.

This phenomenon suggests that the adversarial attack on a source FR model inclines to destroy its decisive and auxiliary features (preferable facial features), yet inevitably overfitting the adversarial examples to the model, and thus degrading the adversarial transferability. Considering this drawback, a seemingly possible way to enhance adversarial transferability is destroying the facial features that are assigned high visual importance by target FR models. However, it is infeasible to directly infer which facial features should be destroyed as to target models when crafting adversarial examples on a source model. On the contrary, if all potentially critical facial features are assigned high visual importance on a source model, one would largely obtain more transferable adversarial examples. Thereby, the critical issue becomes how to spread the visual importance to other potentially critical features in face images on a source model. Considering that the facial features are orthogonal to each other in the pixel space, we formulate this issue into two steps: \textbf{attention-transferring} and \textbf{attention-aggregating}. The attention-transferring is anticipated to transfer the visual importance from the preferable facial features to other potentially critical facial features, aiming to imitating the attentions (characteristics) of target models on the clean face images. Then, attention-aggregation overlaps all transferred visual importance which are ideally orthogonal to each other, ensuring all potentially critical features are to be sufficiently destroyed. 

\begin{algorithm}[t]
\caption{Attention-aggregated Attack}\label{alg:alg1}
\begin{algorithmic}
\STATE \textbf{Input:} A surrogate model $f_\theta$, a benign face image $x^i$ and its reference image $x^r$, maximum perturbation $\epsilon$, number of iteration $T$, momentum $\mu$, step size $\alpha$ and $\beta$, the budget of adversarial noise $\gamma$ and number of iterations $N$ in the first stage of our method.
\STATE \textbf{Output:} An adversarial example $x^{adv}\in{\mathcal{B} _\epsilon(x^i) } $
\STATE \textbf{Initialization:} $x^{adv}_{k(k=0)}=x^{adv}_{t(t=0)}=x^i,g_0=0,\delta_0=0,I=0$
\STATE \textbf{For \(k \) \(\leftarrow\)  \(0\)  \(\textbf{to}\)  \(N-1\)  \(\textbf{do}\):}
\STATE \hspace{0.5cm} Calculate $\delta_{k+1}$ with Equation (\ref{MIM2})
\STATE \hspace{0.5cm} Update $x^{adv}_{k+1}$ with Equation (\ref{MI-update2}) 
\STATE \textbf{End For}
\STATE Calculate $I$ with Equation (\ref{I})
\STATE \textbf{For \(t \) \(\leftarrow\)  \(0\)  \(\textbf{to}\)  \(T-1\)  \(\textbf{do}\):}
\STATE \hspace{0.5cm} $L(x^{adv}_t)=\sum (h^{adv}_t\odot I)$
\STATE \hspace{0.5cm} $g_{t+1}=\mu \cdot g_t +\frac{\bigtriangledown _{x^{adv}_t}L(x^{adv}_t)}{\left \| \bigtriangledown _{x^{adv}_t}L(x^{adv}_t) \right \|_1 } $
\STATE \hspace{0.5cm} $x^{adv}_{t+1}=clip_{(0,\epsilon)}(x^{adv}_{t}+\alpha \cdot sign(g_{t+1}))$
\STATE \textbf{End For}
\STATE \textbf{Return $x^{adv}=x^{adv}_{T}$}
\end{algorithmic}
\label{alg1}

\end{algorithm}

To obtain credible transferred attentions (visual importance), the tranferred attentions should be obtained from the face images whose difference from the clean face image is as least as possible, as the visual importance corresponding to incomplete face images, such as landmark-occluded face images in LGC, is dubious and less transferred. The proposed AAA adopts the adversarial noise, which can maximally transfers the attention within an imperceptible perturbation to the clean images. As depicted in Fig. \ref{attention},  the attention of Resnet50 \cite{Resnet} and Mobilenet \cite{mobilenetv2} (second and third rows) smoothly transfers with iteratively added adversarial perturbation crafted by an off-the-shelf method, such as MIM, which can be regarded as collecting attentions from different FR models (the first row in Fig. \ref{attention}). Specifically, the adversarial face images yielding the transferring attention in Fig. \ref{attention} (second and third rows) are iterated through MIM as follows:
\begin{equation}
    \delta_{k+1}=\mu \cdot \delta_k +\frac{\bigtriangledown _{x^{adv}_k}\mathcal{D}_{f_\theta}(x^{adv}_k,x^{r})}{\left \| \bigtriangledown _{x^{adv}_k}\mathcal{D}_{f_\theta}(x^{adv}_k,x^{r}) \right \|_1 } 
    \label{MIM2}
\end{equation}
\begin{equation}
    x^{adv}_{k+1}=x^{adv}_{k}+\beta \cdot sign(\delta_{k+1})
    \label{MI-update2}
\end{equation}
where $x^{adv}_0$ is initialized to the clean input face image, $\delta_{k+1}$ denotes the adversarial noise crafted in the $k$-$th$ iteration, which is to transfer the attention of the source model on $x^{adv}_{k}$, and $\beta$ is the step size to generate $x^{adv}_{k+1}$. Then, the attention-aggregating step overlaps all calculated attentions that are ideally orthogonal to each other, to enable the participation of the vast majority of facial features in the attacking process as follows:
\begin{equation}
    I=\sum_{k=0}^{N-1}\frac{\partial \mathcal{D}_{f_\theta}(x^{adv}_k,x^{r})}{h^{adv}_k}  \label{I}
\end{equation}
where $N$ is the number of iterations to generate adversarial face images carrying orthogonal visual importance (or attention) using Equation (\ref{MIM2}) and (\ref{MI-update2}), and $h^{adv}_k$ represents the mid-layer feature maps extracted from $x^{adv}_k$. 

\begin{table*}[!t]
    \begin{center}
    \caption{ \textbf{Attack success rate(\%) of MIM, LGC, AAA, and their DI-combined versions against models built on different backbones and supervised by various losses, in terms of dodging and impersonation. The first column lists source models and the second row lists target models. The best results are highlighted in bold.}}
    \centering
    \vspace{4mm}
    \label{t1}
    \begin{tabular}{cc|cccc|cccc}
    \hline
        \quad & \quad & \multicolumn{4}{c|}{Dodging} & \multicolumn{4}{c}{Impersonation} \\
        \hline
        \quad & Method & \makecell[c]{FaceNet}& \makecell[c]{SphereFace} & \makecell[c]{CosFace} &\makecell[c]{ArcFace} & \makecell[c]{FaceNet}& \makecell[c]{SphereFace} & \makecell[c]{CosFace} &\makecell[c]{ArcFace} \\
    \hline
        \multirow{6}{*}{\makecell[c]{FaceNet}}&MIM& 100 & 70.10 & 62.17 & 45.40&100&44.70&39.83&23.90 \\ 
        &MI-DIM& 99.93 & 75.87 & 68.00 &
        51.93&99.86&49.97&44.93&30.10 \\ 
        &LGC&100 & 71.83 & 63.50 & 46.10&100&46.63&40.57&25.20 \\ 
        &LGC-DI&99.93 & 77.50 & 70.90 &55.27& 99.83&53.80&46.47&34.30 \\ \rowcolor[gray]{0.9}
        &AAA & \textbf{100} & 78.17 & 69.93 &53.17 &\textbf{100} &50.97&45.87&31.90 \\ \rowcolor[gray]{0.9}
        &AAA-DI & 100&\textbf{85.43}&\textbf{81.17}&\textbf{67.73}&99.93&\textbf{58.63}&\textbf{52.50}&\textbf{41.53} \\ 
        \hline
        \multirow{6}{*}{\makecell[c]{SphereFace}}&MIM& 63.20 & 100 & 95.60 & 62.43&27.93&100&81.00&47.53 \\ 
        &MI-DIM& 66.20&99.97&97.57&65.00&31.00&100.0&84.83&50.03 \\ 
        
        &LGC&63.90 & 100 & 96.13 & 63.43&29.87&100&82.83&47.97 \\ 
        &LGC-DI&67.40&99.93&97.13&66.53&33.97&99.93&84.80&51.67 \\ 
        \rowcolor[gray]{0.9}
        &AAA & 67.20 & \textbf{100} & 98.50 & 66.63 &30.90&\textbf{100}&83.57&51.47 \\ \rowcolor[gray]{0.9}

        &AAA-DI& \textbf{74.77}&99.93&\textbf{99.53}&\textbf{68.67}&\textbf{35.57}&100&\textbf{85.40}&\textbf{54.73} \\ 
        \hline
        \multirow{6}{*}{\makecell[c]{CosFace}}&MIM& 61.50 & 92.27 & 100 & 67.17 & 29.53 & 74.23&100&49.67 \\ 
        &MI-DIM&67.80&97.10&99.97&69.73&31.43&76.47&100&53.63 \\ 
        &LGC&65.00 & 97.87 & 100 & 68.23 & 29.90 & 74.93&100&50.53  \\ 
        &LGC-DI&67.87&98.00&99.93&71.50&32.17&76.87&99.97&53.57  \\ 
        \rowcolor[gray]{0.9}
        &AAA & 68.60 & 97.93 & \textbf{100} & 71.53 & 33.77&77.87&\textbf{100}&55.53 \\ \rowcolor[gray]{0.9}

         &AAA-DI & \textbf{70.27}&\textbf{98.83}&99.20&\textbf{77.87}&\textbf{37.37}&\textbf{82.70}&99.63&\textbf{59.97} \\ 
        \hline
        \multirow{6}{*}{\makecell[c]{ArcFace}}&MIM& 61.63 & 82.17 & 78.67 & 100 & 29.93 & 67.47&63.40&100\\ 
        &MI-DIM& 66.50&86.93&83.97&100&33.13&72.40&67.73&100\\ 
        
        &LGC&63.63 & 83.40 & 79.73 & 100 & 31.83 & 69.80&65.00&100  \\ 
        &LGC-DI&67.53&87.40&84.80&99.97&34.63&73.27&67.83&100  \\ 
        \rowcolor[gray]{0.9}
        &AAA & 69.10 & 86.50 & 83.90 & 100 & 34.63&71.87&67.80&100 \\ 
        \rowcolor[gray]{0.9}
        &AAA-DI & \textbf{73.50}&\textbf{90.47}&\textbf{87.67}&\textbf{100}&\textbf{37.37}&\textbf{75.40}&\textbf{70.97}&\textbf{100} \\ 
        \hline
        
    \end{tabular}
    \end{center}
\end{table*}

\begin{table*}[!t]
    \begin{center}
    \caption{ \textbf{Attack success rate(\%) of MIM, LGC, AAA, and their DI-combined versions against models built on different backbones and supervised by LMCL loss, in terms of dodging and impersonation. The first column lists source models and the second row lists target models. The best results are highlighted in bold.}}
    \centering
     \vspace{4mm}
    \label{t2}
    \begin{tabular}{cc|cccc|cccc}
    \hline
        \quad & \quad & \multicolumn{4}{c|}{Dodging} & \multicolumn{4}{c}{Impersonation} \\
        \hline
        \quad & Method & \makecell[c]{Mobile\\Face} &  \makecell[c]{Mobile\\Net-V2} & \makecell[c]{Shuffle\\Net-V1} &\makecell[c]{ResNet50} & \makecell[c]{Mobile\\Face} &  \makecell[c]{Mobile\\Net-V2} & \makecell[c]{Shuffle\\Net-V1} &\makecell[c]{ResNet50} \\
    \hline
        \multirow{6}{*}{\makecell[c]{MobileFace}}
        &MIM& 100 & 89.53 & 93.43 & 89.40 & 100 & 78.47&96.33&82.33 \\ 
        &MI-DIM& 100&91.50&95.40&93.13&100&80.57&97.53&84.70 \\
        &LGC&100 & 90.63 & 94.13 & 89.93 & 100 & 80.10&96.80&83.13  \\ 
        &LGC-DI&100&92.03&95.73&94.00&100&82.80&97.10&85.33  \\
        \rowcolor[gray]{0.9}
        &AAA & 100 & 94.17 & 96.00 & 93.43 & 100&83.73&97.93&86.83 \\
        \rowcolor[gray]{0.9}
        &AAA-DI & \textbf{100}&\textbf{95.37}&\textbf{98.00}&\textbf{96.97}&\textbf{100}&\textbf{85.27}&\textbf{98.77}&\textbf{89.03} \\
        \hline
        \multirow{6}{*}{\makecell[c]{MobileNet-V2}}&MIM& 83.03 & 100 & 71.23 & 64.03 &75.40 & 100&71.67&48.03 \\ 
        &MI-DIM& 84.40&99.97&79.53&70.13&75.20&100&76.07&51.47 \\ 
        
        &LGC&83.93 & 100 & 73.20 & 65.87 & 77.60 &100&73.10&48.63 \\ 
        &LGC-DI&85.77&99.90&77.17&71.47&79.27&100&77.03&51.67 \\ 
        \rowcolor[gray]{0.9}
        &AAA & 88.00 & 100 & 77.23 &71.83 & 81.37&100&77.97&55.67 \\ 
        \rowcolor[gray]{0.9}
        &AAA-DI & \textbf{91.03}&\textbf{100}&\textbf{88.37}&\textbf{83.10}&\textbf{84.17}&\textbf{100}&\textbf{81.70}&\textbf{57.77} \\ 
        \hline
        \multirow{6}{*}{\makecell[c]{ShuffleNet-V1}}&MIM& 90.43 & 76.77 & 100 & 75.90 & 93.33 & 65.43&100&71.73 \\ 
        &MI-DIM& 93.63&83.50&99.93&80.80&94.87&69.97&100&76.97 \\ 
        &LGC&90.77 & 78.37 & 100 & 77.07 & 93.80 & 66.57&100&72.00  \\ 
        &LGC-DI&92.20&85.20&99.97&81.97&95.00&70.77&100&77.10 \\ 
        \rowcolor[gray]{0.9}
        &AAA & 94.23 & 83.27 & \textbf{100} & 82.67 & 96.30&72.57&\textbf{100}& 78.47\\ \rowcolor[gray]{0.9}
        &AAA-DI & \textbf{95.73}&\textbf{90.27}&99.97&\textbf{88.43}&\textbf{97.33}&\textbf{75.70}&100&\textbf{83.73}\\ 
        \hline
        \multirow{6}{*}{\makecell[c]{ResNet50}}&MIM& 93.57 & 83.33 & 87.87 & 100 & 88.70 & 60.27 &87.83&100\\ 
        &MI-DIM& 95.53&88.40&89.80&99.93&90.03&62.07&89.23&100\\ 
        
        &LGC&93.67 & 85.33 & 88.30 & 100 & 87.43 & 61.17&87.97&100  \\ 
        &LGC-DI&95.27&88.63&90.30&99.83&90.07&64.50&89.83&99.97  \\ 
        \rowcolor[gray]{0.9}
        &AAA & 97.67 & 90.40 & 93.57 & \textbf{100} & 91.57&66.00&91.17&\textbf{100} \\ \rowcolor[gray]{0.9}
        &AAA-DI & \textbf{98.60}&\textbf{93.93}&\textbf{96.00}&99.97&\textbf{94.33}&\textbf{69.17}&\textbf{93.27}&100 \\ 
        \hline
        
    \end{tabular}
    \end{center}
    \vspace{3mm}
\end{table*}

\begin{table*}[t]
    \begin{center}
    \caption{\textbf{Attack success rate(\%) of MIM, LGC, AAA, and their DI-combined versions against models built on IResNet50 and supervised by various loss, in terms of dodging and impersonation. The first column lists source models and the second row lists target models. The best results are highlighted in bold.}}
    \centering
     \vspace{4mm}
    \label{t3}
    \begin{tabular}{cc|cccc|cccc}
    \hline
        \quad & \quad & \multicolumn{4}{c|}{Dodging} & \multicolumn{4}{c}{Impersonation} \\
        \hline
        \quad & Method & \makecell[c]{SoftMax\\-IR} &  \makecell[c]{Sphere\\Face-IR} & \makecell[c]{CosFace\\-IR} &\makecell[c]{ArcFace\\-IR} & \makecell[c]{SoftMax\\-IR} &  \makecell[c]{Sphere\\Face-IR} & \makecell[c]{CosFace\\-IR} &\makecell[c]{ArcFace\\-IR}   \\
    \hline
        \multirow{6}{*}{\makecell[c]{SoftMax-IR}}&MIM& 100 & 98.77 & 94.23 & 97.27 & 100 & 92.73&86.40&91.87 \\ 
        &MI-DIM&100&99.27&95.17&98.97&100&94.43&87.17&92.23 \\ 
        
        &LGC&100 & 98.87 & 94.27 & 97.70 & 100 & 92.50&86.00&92.07  \\ 
        &LGC-DI&99.97&98.97&96.03&98.93&99.97&94.00&88.73&94.43 \\ \rowcolor[gray]{0.9}
        
        &AAA & \textbf{100} & 99.67 & 99.00 & 99.07 & \textbf{100}&95.33&89.27&94.03 \\ \rowcolor[gray]{0.9}
        &AAA-DI &100&\textbf{99.70}&\textbf{99.27}&\textbf{99.57}&100&\textbf{97.57}&\textbf{92.90}&\textbf{96.50} \\ 
        \hline
        \multirow{6}{*}{\makecell[c]{SphereFace-IR}}&MIM& 97.23& 100 & 90.07 & 95.20 & 83.30 & 100&66.07&75.10 \\ 
        &MI-DIM& 98.90&100&92.30&96.50&85.93&100&66.63&76.47 \\
        
        &LGC&97.57 & 100 & 90.13 & 95.23 & 84.00 & 100&67.67&76.87 \\ 
        &LGC-DI&98.30&100&92.67&97.20&87.30&100&68.37&78.17 \\ 
        \rowcolor[gray]{0.9}
        
        &AAA & 99.20 & \textbf{100} & 95.90 & 98.23 & 88.13&\textbf{100}&72.33&81.17 \\ \rowcolor[gray]{0.9}
         &AAA-DI &\textbf{99.43}&100&\textbf{97.67}&\textbf{98.93}&\textbf{90.60}&100&\textbf{76.00}&\textbf{85.03} \\ 
        \hline
        \multirow{6}{*}{\makecell[c]{CosFace-IR}}&MIM& 94.13 & 94.40 & 100 & 99.70 & 62.50 & 56.53&100&99.47 \\ 
        &MI-DIM&95.70&94.67&100&99.60&65.70&59.33&100&99.27 \\
        
        &LGC&95.30 & 95.13 & 100 & 99.70& 63.10 & 56.33&100&99.47  \\ 
        &LGC-DI&97.07&97.10&100&99.83&66.27&59.73&100&99.03  \\ 
        \rowcolor[gray]{0.9}
        &AAA & 98.07 & 97.87 & \textbf{100} & 98.83 & 67.33&60.00&\textbf{100}&99.70 \\ \rowcolor[gray]{0.9}
         &AAA-DI & \textbf{98.40}&\textbf{98.50}&100&\textbf{99.87}&\textbf{71.50}&\textbf{64.80}&100&\textbf{99.93} \\ 
        \hline
        \multirow{6}{*}{\makecell[c]{ArcFace-IR}}&MIM& 96.53 & 96.37 & 99.63 & 100 & 81.53 & 74.07&99.87&100 \\ 
        &MI-DIM& 97.67&96.70&99.20&100&84.93&76.00&99.73&100 \\
        
        &LGC&96.93 & 97.30 & 99.67& 100 & 81.83 & 75.03&99.87&100  \\ 
        &LGC-DI&97.33&98.70&98.97&100&83.13&77.50&99.83&100 \\ 
        \rowcolor[gray]{0.9}
        &AAA & 99.77 & 98.83 & 99.27 & \textbf{100} & 86.10&79.40&99.27&\textbf{100} \\ \rowcolor[gray]{0.9}
        &AAA-DI & \textbf{99.93}&\textbf{99.53}&\textbf{99.70}&100&\textbf{89.17}&\textbf{83.47}&\textbf{99.87}&100 \\ 
        \hline
        
    \end{tabular}
    \end{center}
\end{table*}

\begin{table*}[t]
    \begin{center}
    \caption{\textbf{Attack success rate(\%) of MIM, LGC, AAA, and their increasingly complex combined versions when $\epsilon$ is set to 4.0 and 8.0, respectively. All adversarial examples are crafted on MobileFace, in terms of dodging and impersonation. The second column lists attack methods and the second row lists target models. The best results are highlighted in bold.}}
    \centering
     \vspace{4mm}
    \label{t4}
    \begin{tabular}{c|c|cccc|cccc}
    \hline
        \quad & \quad & \multicolumn{4}{c|}{Dodging} & \multicolumn{4}{c}{Impersonation} \\
        \hline
       \quad & Method & \makecell[c]{Mobile\\Face} &  \makecell[c]{Mobile\\Net-V2} & \makecell[c]{Shuffle\\Net-V1} &\makecell[c]{ResNet50} & \makecell[c]{Mobile\\Face} &  \makecell[c]{Mobile\\Net-V2} & \makecell[c]{Shuffle\\Net-V1} &\makecell[c]{ResNet50} \\
    \hline
        \multirow{13}{*}{$\epsilon$=4.0}
        &MIM& 99.17&40.40&44.23&31.27&100&37.97&70.47&42.57 \\ 
        &LGC&99.23&41.40&45.43&32.07&100&38.57&71.73&43.20  \\
         &AAA & \textbf{99.80}&\textbf{45.60}&\textbf{50.13}&\textbf{36.67}&\textbf{100}&\textbf{42.20}&\textbf{74.43}&\textbf{46.10} \\
         \cmidrule(r){2-10} 
        &MI-DIM&91.97&45.10&51.63&38.47&99.87&41.63&75.30&46.83 \\ 
         &LGC-DI&93.90&46.83&53.13&40.27&99.87&43.40&76.90&48.53 \\ 
          &AAA-DI &\textbf{99.47}&\textbf{52.60}&\textbf{58.60}&\textbf{45.87}&\textbf{100}&\textbf{47.83}&\textbf{80.80}&\textbf{53.47} \\
          \cmidrule(r){2-10} 
        &MI-DI-SIM&92.17&52.57&57.60&45.90&99.83&44.67&77.83&49.47 \\ 
        &LGC-DI-SIM&92.97&54.07&58.17&46.63&99.77&46.60&77.97&50.77 \\
         &AAA-DI-SIM&\textbf{99.43}&\textbf{56.47}&\textbf{61.53}&\textbf{49.60}&\textbf{100}&\textbf{51.67}&\textbf{82.73}&\textbf{56.80} \\ 
         \cmidrule(r){2-10} 
       
        &MI-DI-SIM-SG&92.83&53.77&59.17&47.33&99.87&49.30&81.03&53.73 \\  
        &LGC-DI-SIM-SG&92.30&55.47&60.13&50.10&99.87&50.23&82.97&55.40 \\
       
        &AAA-DI-SIM-SG&\textbf{99.43}&\textbf{58.60}&\textbf{63.27}&\textbf{55.03}&\textbf{100}&\textbf{56.33}&\textbf{86.17}&\textbf{62.40} \\ 
       
         \cmidrule(r){1-10} 
        \multirow{13}{*}{$\epsilon$=8.0}
        &MIM& 100 & 89.53 & 93.43 & 89.40 & 100 & 78.47&96.33&82.33 \\ 
        &LGC&100 & 90.63 & 94.13 & 89.93 & 100 & 80.10&96.80&83.13  \\ 
        &AAA & \textbf{100} & \textbf{94.17} & \textbf{96.00} & \textbf{93.43} & \textbf{100}&\textbf{83.73}&\textbf{97.93}&\textbf{86.83} \\
        \cmidrule(r){2-10} 
        &MI-DIM& 100&91.50&95.40&93.13&100&80.57&97.53&84.70 \\
        &LGC-DI&100&92.03&95.73&94.00&100&82.80&97.10&85.33  \\
        &AAA-DI & \textbf{100}&\textbf{95.37}&\textbf{98.00}&\textbf{96.97}&\textbf{100}&\textbf{85.27}&\textbf{98.77}&\textbf{89.03} \\
          \cmidrule(r){2-10} 
        &MI-DI-SIM&100&93.80&96.20&95.00&100&83.83&98.70&86.90 \\ 
         &LGC-DI-SIM&100&94.67&97.93&96.57&100&84.80&98.40&87.93 \\ 
          &AAA-DI-SIM&\textbf{100}&\textbf{96.83}&\textbf{98.97}&\textbf{97.83}&\textbf{100}&\textbf{88.37}&\textbf{99.10}&\textbf{91.10} \\ 
        \cmidrule(r){2-10} 
        &MI-DI-SIM-SG&100&95.57&97.93&97.90&100&85.47&99.23&89.57 \\ 
         
        &LGC-DI-SIM-SG&100&96.23&98.37&97.70&100&87.13&99.23&90.60 \\
      
        &AAA-DI-SIM-SG&\textbf{100}&\textbf{97.03}&\textbf{98.97}&\textbf{98.07}&\textbf{100}&\textbf{94.67}&\textbf{99.87}&\textbf{96.40} \\
        \hline
        
    \end{tabular}
    \end{center}
\end{table*}

From the visual perspective, aggregating the attentions listed in the second or third row in Fig. \ref{attention} can be regarded as imitating to aggregating the attentions obtained from various FR models listed in the first row in Fig. \ref{attention}, achieving a special kind of ensemble-like effect. Since the aggregation of attentions listed in one of these three rows will spread the visual importance over the holistic face while $N$ tends toward $\infty$, our objective of assigning visual importance to all potentially critical features on a source model can be reached via the attention-aggregating step.

Then, substituting $I$ into Equation (\ref{feature_attack}), we finally obtain the attacking loss of our method:
\begin{equation}
    L(x^{adv})=\sum \left (  \sum_{k=0}^{N-1}\frac{\partial \mathcal{D}_{f_\theta}(x^{adv}_k,x^{r})}{h^{adv}_k}\right )\odot h^{adv}
    \label{ourloss}
\end{equation}
where $h^{adv}$ denotes feature maps extracted from the desired adversarial example $x^{adv}$. For the sake of fairness, the popular momentum-based iterating scheme is applied to minimize (\ref{ourloss}), as summarized in Algorithm \ref{alg:alg1}.

To investigate the possible rationality supporting the strong adversarial transferability in our method, the gradient of the attacking loss w.r.t. the input image can be written as:
\begin{align}
\frac{\partial L(x^{adv}_t)}{\partial x^{adv}_t}  &=\sum  \left (  \sum_{k=0}^{N-1}\frac{\partial \mathcal{D}_{f_\theta}(x^{adv}_k,x^{r})}{h^{adv}_k}\right )\odot \frac{\partial h^{adv}_t}{\partial x^{adv}_t}
    \nonumber
    \\&=\sum_{i=0}^{m-1} \left (  \sum_{k=0}^{N-1}\frac{\partial \mathcal{D}_{f_\theta}(x^{adv}_k,x^{r})}{(h^{adv}_k)_i}\right )\cdot \frac{\partial (h^{adv}_t)_i}{\partial x^{adv}_t}
\end{align}
where $m$ denotes the number of activation units constructing the feature maps. 
For an intuitive comparison, $\bigtriangledown _{x^{adv}_t}\mathcal{D}_{f_\theta}(x^{adv}_t,x^{r})$ in (\ref{MI-FGSM}) can be decomposed as:    
\begin{equation}
    \bigtriangledown _{x^{adv}_t}\mathcal{D}_{f_\theta}(x^{adv}_t,x^{r})=\sum_{i=0}^{m-1} \frac{\partial \mathcal{D}_{f_\theta}(x^{adv}_t,x^{r})}{\partial (h^{adv}_t)_i} \cdot  \frac{\partial (h^{adv}_t)_i}{\partial x^{adv}_t}
\end{equation}

Apparently, the unique difference between MIM and our method is that the dynamic gradients $\frac{\partial \mathcal{D}_{f_\theta}(x^{adv}_t,x^{r})}{\partial h^{adv}_t}$ varying with $t$ is found working in MIM, while the static gradients $\sum_{k=0}^{N-1}\frac{\partial \mathcal{D}_{f_\theta}(x^{adv}_k,x^{r})}{h^{adv}_k}$ conduct the feature-level attack in our method. Fig. \ref{attention} illustrates the intuition of our method. Taking the attention of ResNet50 in Fig. \ref{attention} as an example, the attention smoothly transfers with iteratively added adversarial noise, which can be regarded as getting attention from different FR models (the first row in Fig. \ref{attention}). Furthermore, the transferring attentions listed in the second and third rows in Fig. \ref{attention} indicates that MIM has also realized the objective of attacking all potentially critical facial features. However, the unique point differing from our method is that the facial features emerging in the $t$-$th$ attacking iteration are not necessarily to emerge and attend to be attacked in the $(t+1)$-$th$ iteration when generating adversarial examples using MIM as shown in the second and third rows in Fig. \ref{attention}, revealing that MIM is unable to continuously attack holistic facial features and thus resulting in inadequate destruction and weak adversarial examples. As to our method, sufficient destruction to all potentially critical facial features is guaranteed via the attention-transferring and attention-aggregating step, contributing to stronger adversarial examples.

\begin{table*}[t]
    \begin{center}
    \caption{\textbf{Attack success rate(\%) of MIM, LGC, AAA, and their DI-combined version when $\epsilon$ is set to 4.0. All adversarial examples are crafted on an ensemble of models including MobileFace, SoftMax-IR, and ArcFace, in terms of dodging. The first column lists attack methods and the first row lists target models. The best results are highlighted in bold.}}
    \centering
     \vspace{3mm}
    \label{t5}
    \begin{tabular}{cccccccccc}
    \hline
       
       Method & \makecell[c]{Mobile\\Face} & \makecell[c]{SoftMax\\-IR} &  \makecell[c]{ArcFace} & \makecell[c]{Mobile\\Net-V2} &\makecell[c]{Shuffle\\Net-V1} &\makecell[c]{Sphere\\Face} & \makecell[c]{CosFace} &\makecell[c]{Sphere\\Face-IR} & \makecell[c]{CosFace\\-IR} \\
    \hline
        MIM& 95.50&90.13&93.63&57.20&58.10&55.47&46.07&66.17&61.20 \\ 
        MI-DIM&96.97&93.63&94.87&61.63&61.30&59.53&50.47&69.90&64.37 \\ 
       
        LGC&95.30&90.33&93.60&58.83&59.33&56.63&47.27&66.93&61.63  \\ 
        LGC-DI&96.83&93.97&95.27&61.70&62.47&59.59&49.63&69.43&64.87 \\ 
      \rowcolor[gray]{0.9}
        AAA & 96.93&94.73&96.07&62.60&63.03&59.80&49.77&71.97&68.07 \\ 
        \rowcolor[gray]{0.9}
        AAA-DI &\textbf{97.40}&\textbf{96.33}&\textbf{97.97}&\textbf{68.33}&\textbf{68.50}&\textbf{65.47}&\textbf{55.63}&\textbf{77.30}&\textbf{74.17} \\

        \hline

    \end{tabular}
    \end{center}
\end{table*}

\begin{table*}[t]
    \begin{center}
    \caption{\textbf{Attack success rate(\%) of MIM, LGC, AAA, and their DI-combined version when $\epsilon$ is set to 4.0. All adversarial examples are crafted on an ensemble of models including MobileFace, SoftMax-IR, and ArcFace, in terms of impersonation. The first column lists attack methods and the first row lists target models. The best results are highlighted in bold.}}
    \centering
     \vspace{3mm}
    \label{t6}
    \begin{tabular}{cccccccccc}
    \hline
       
       Method & \makecell[c]{Mobile\\Face} & \makecell[c]{SoftMax\\-IR} &  \makecell[c]{ArcFace} & \makecell[c]{Mobile\\Net-V2} &\makecell[c]{Shuffle\\Net-V1} &\makecell[c]{Sphere\\Face} & \makecell[c]{CosFace} &\makecell[c]{Sphere\\Face-IR} & \makecell[c]{CosFace\\-IR} \\
    \hline
        MIM& 99.97&95.07&99.43&57.07&77.60&48.57&44.67&61.07&81.70 \\ 
        MI-DIM&99.80&97.07&99.67&60.13&79.93&52.27&49.80&64.63&84.73 \\ 
       
        LGC&99.83&94.33&99.27&58.33&77.70&49.37&45.20&61.23&81.53  \\ 
        LGC-DI&99.87&97.13&99.73&61.50&79.27&53.63&50.83&64.13&84.50 \\ 
        \rowcolor[gray]{0.9}
        AAA & 99.77&97.63&99.47&62.73&80.83&53.90&49.60&65.27&84.57 \\ 
        \rowcolor[gray]{0.9}
        AAA-DI &\textbf{99.03}&\textbf{98.73}&\textbf{99.83}&\textbf{68.73}&\textbf{88.77}&\textbf{60.63}&\textbf{55.63}&\textbf{69.50}&\textbf{89.83} \\

        \hline

    \end{tabular}
    \end{center}
\end{table*}

\section{Experimental Results}
In this section, three groups of comparative experiments are conducted to verify the effectiveness of our method. In addition, we combined our method with existing universal boosting methods to further enhance the adversarial transferability. Besides, we arranged the experiment of attacking an ensemble of models to further validate the effectiveness and superiority of the proposed method. Finally, ablation experiments are conducted to probe the influence of two hyper-parameters, including the choice of the layer to attack and the budget of adversarial noise to transfer the attention of a source model.

\subsection{Experimental Settings}
\noindent\textbf{Datasets.} Following the previous work \cite{LGC}, we take the datasets including 3000 pairs of face images randomly sampled from the LFW \cite{LFW} benchmarking datasets.  

\vspace{\baselineskip}
\noindent\textbf{Models.} We select leading FR models including  FaceNet \cite{FaceNet}, SphereFace \cite{sphereface}, CosFace \cite{cosface}, ArcFace \cite{deng2019arcface}, MobileFace \cite{MobileFace},  MobileNetV2 \cite{mobilenetv2}, ShuffleNet-V1 \cite{shufflenetv1}, ResNet50 \cite{Resnet},  Softmax-IR \cite{deng2019arcface}, SphereFace-IR, CosFace-IR, and ArcFace-IR. These models are classified into three groups according to their backbones and training losses. 

\vspace{\baselineskip}
\noindent\textbf{Attacking Methods.} We take MIM as our baseline, whose destruction to the facial features is insufficient compared to our method according to the analysis in section \ref{AAA}. We also select LGC, which is also inspired by transferring the attention of a surrogate model, as a competitor. However, we believe that our method transfers the attention to a broader spectrum than LGC in that the visual importance corresponding
to incomplete face images, such as landmark-occluded face images in LGC, is dubious and less transferred. 

\vspace{\baselineskip}
\noindent\textbf{Parameters.} To craft unnoticeable noise and enhance the attacking efficiency, we set the maximum perturbation to 8/255, the number of iterations to 10, the step size to 1.2/255, and the momentum to 1.0 for all methods. For LGC, the mask officially consists of four squares with a side length of 7 pixels randomly sampled from all prior facial landmarks. For our method, the budget of adversarial noise $\gamma$ to spread the visual importance from preferable facial features is set to 8/255. As for the layer to attack, we choose repeat.2\_5 in FaceNet, conv3\_3 in SphereFace, layer.3 in CosFace, blocks.7 in MobileFace, features.9 in MobileNetV2, stage3.ShuffleUnit\_Stage3\_3 in ShuffleNet-V1, layer2.3 for ResNet50, and body.19 for ArcFace, Softmax-IR, SphereFace-IR, CosFace-IR, and ArcFace-IR.

\subsection{Attacking Results}
\subsubsection{Attacking A Single FR Model}

Involved FR models are divided into three groups according to different selections on backbones and training losses, to verify the robust effectiveness and superiority of our method. First, models with neither same backbones nor same training losses belong to the first group to mimic real-world situations. Second, models that are built on different backbones and supervised by the same training loss, e.g., LMCL, constitute the second group. Third, models that are built on the same backbone, e.g., IResNet50 \cite{deng2019arcface} and supervised by different training losses constitute the third group. We employed a simple and effective data augmentation method, e.g. input diversity (DI-FGSM) \cite{DIM}, which is motivated by the idea that
adversarial examples lose their maliciousness after simple
transformations, to first enhance the black-box attacking performance. Table \ref{t1} - \ref{t3} illustrate the experimental results of these three groups of models when attacking a single model, which show that our method outperforms other two methods on the black-box attack whether combining with the input diversity \cite{DIM} or not, while maintaining the strong attacking intensity to white-box models as the ASRs against source models consistently remain 100\%. These three groups of experimental results suggest that the proposed method is robust and stable when attacking various kinds of FR models, thus validating the superiority of our method over the baseline methods.

\subsubsection{Combination with Other Methods}
Considering that the proposed method can be enhanced by combining with the DI-FGSM \cite{DIM} according to Table \ref{t1} - \ref{t3}, it can be further improved by integrating with other boosting methods, such as SI-FGSM (SIM) \cite{NI-SIM}, SmoothGrad (SG) \cite{smoothgrad} or the combination of them. First, SI-FGSM \cite{NI-SIM} introduces the concept of loss-preserving transformations. That is, if a
transformed image maintains the model’s predicted label to the
original image, this transformation can be regarded as a loss-preserving
transformation. SI-FGSM observes that changes in luminance
do not alter the label but only slightly decrease true label
confidence. This insight is utilized for generating additional
samples for improving adversarial transferability. Second, given the ReLU activation
function, networks are non-continuously differentiable at the
units that are not activated. Indeed, the non-linear nature of
networks results in unstable gradients with rapid fluctuations. To mitigate this instability, SmoothGrad (SG) \cite{smoothgrad} proposes that the gradient at any given point is less significant than the mean gradients of the neighborhoods, which is seemingly to be utilized to generate additional
samples to improve adversarial transferability. Since these three boosting methods are independent with each other, they can be simultaneously applied to boost the attacking performance of the baselines and our method. Table \ref{t4} records the experimental results of the increasingly complex attack methods based on the baselines and our method when the maximum perturbation $\epsilon$ is set to 4.0 and 8.0, respectively. It is found that the ASRs are coherently improved with the increasingly complex combination. It also shows that the proposed method consistently outperforms the LGC \cite{LGC} and MIM \cite{MIM} when they are combined with the DIM, DI-SIM (simultaneously combined with DIM and SIM), and DI-SIM-SG (simultaneously combined with DIM, SIM and SG). The superiority of AAA may come from two hands. First, the attention-transferring in AAA diverts the visual importance to imitate the attentions (characteristics) of target models on
the clean face images while MIM drives the perturbation to merely destroy the decisive and auxiliary facial features for the source model in the face images. LGC occludes the facial landmarks while damaging the integrity of the face image, probably generating dubious and less transferred visual importance. Second, the attention-aggregation in AAA ensures all potentially critical features are to be sufficiently attacked while MIM is unable to continuously attack all extracted facial features and LGC is also confronted with such an issue, thus resulting in inadequate destruction and weak adversarial example.

\subsubsection{Attacking An Ensemble of FR Models}
Since the attacking performance is relatively unsatisfactory when the maximum perturbation $\epsilon$ is set to 4.0 according to Table \ref{t4}, we try to further improve the adversarial transferability by crafting adversarial examples using an ensemble of FR models, which explicitly brings the perturbation fooling multiple models into adversarial examples, aiming at preventing it from overfiting to the models built on the similar backbone (for example, adversarial examples crafted on Mobilenet \cite{mobilenetv2} are very likely to fool Mobilenet-V2, which are however limited to fool models built on other backbones, such as SphereFace \cite{sphereface} or ResNet \cite{resnet52}). Table \ref{t5} - \ref{t6} exhibit the experimental results of attacking an ensemble of models built on different backbones, including MobileFace, SoftMax-IR, and ArcFace. The results show that all methods are improved compared to Table \ref{t4} and the proposed method still outperforms the baselines.

\begin{figure*}[t]
    \centering
    \subfigure[ \textbf{Dodging}]{
        \includegraphics[width=0.48\linewidth,height=0.35\linewidth]{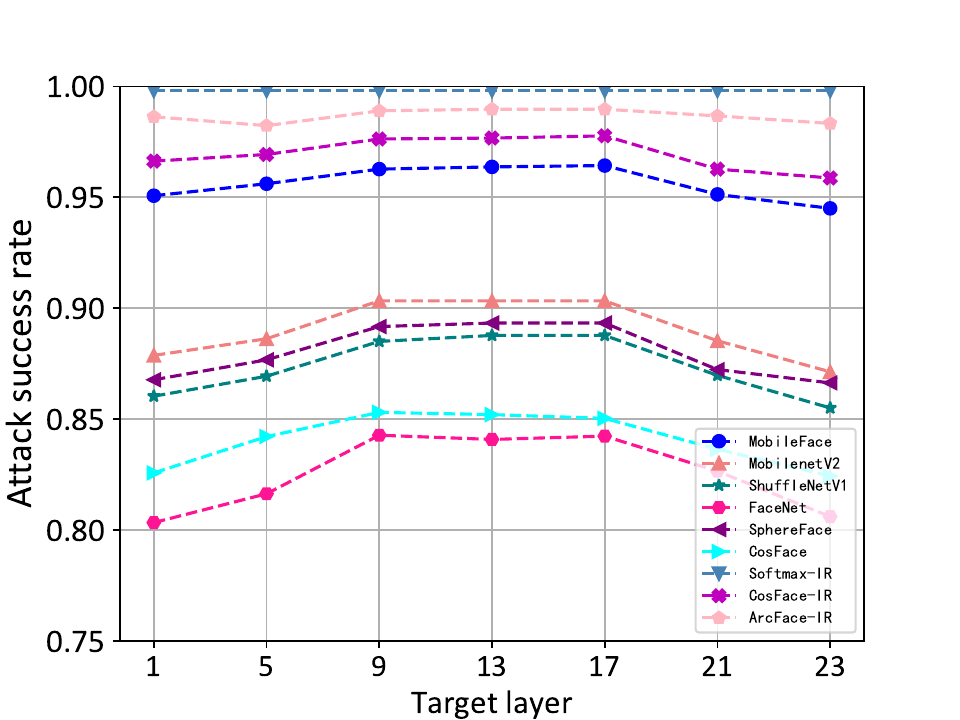}}~~\hfill~
    \subfigure[\textbf{Impersonation}]{
        \includegraphics[width=0.48\linewidth,height=0.35\linewidth]{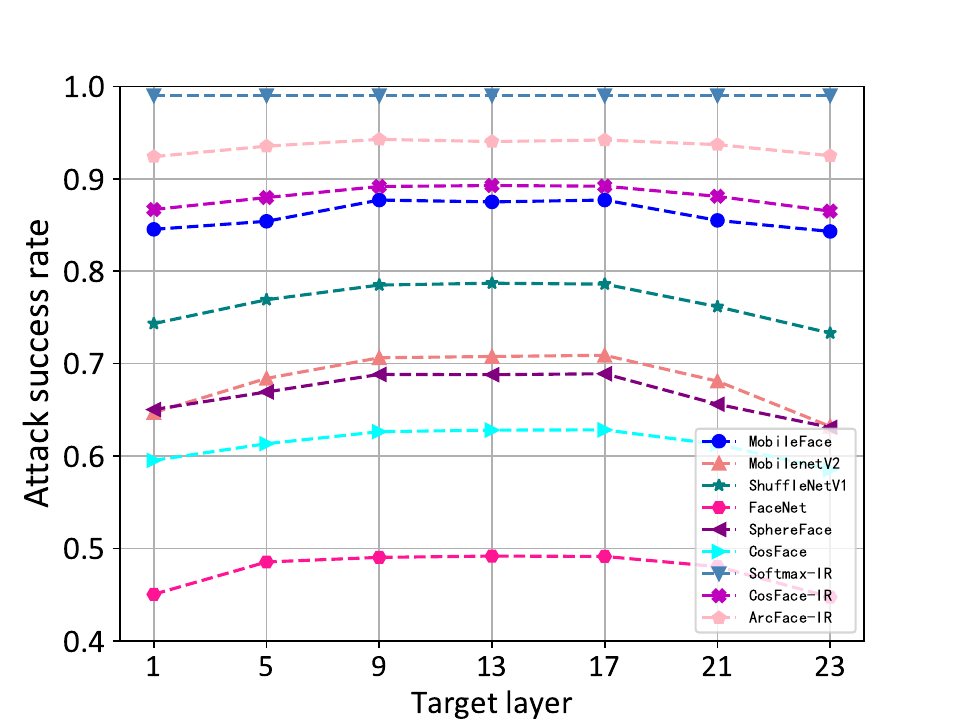}}~~\hfill~

    \caption{ \textbf{The influence of layer choice on ASR. Shallow, middle, and deep layers are respectively attacked in Softmax-IR to craft adversarial examples with our method whose success rates against different target models are exhibited.}}
    \label{layers_ablation}
    \vspace{5mm}
\end{figure*}

\begin{figure*}[!t]
    \centering
    \subfigure[ \textbf{Dodging}]{
        \includegraphics[width=0.48\linewidth,height=0.35\linewidth]{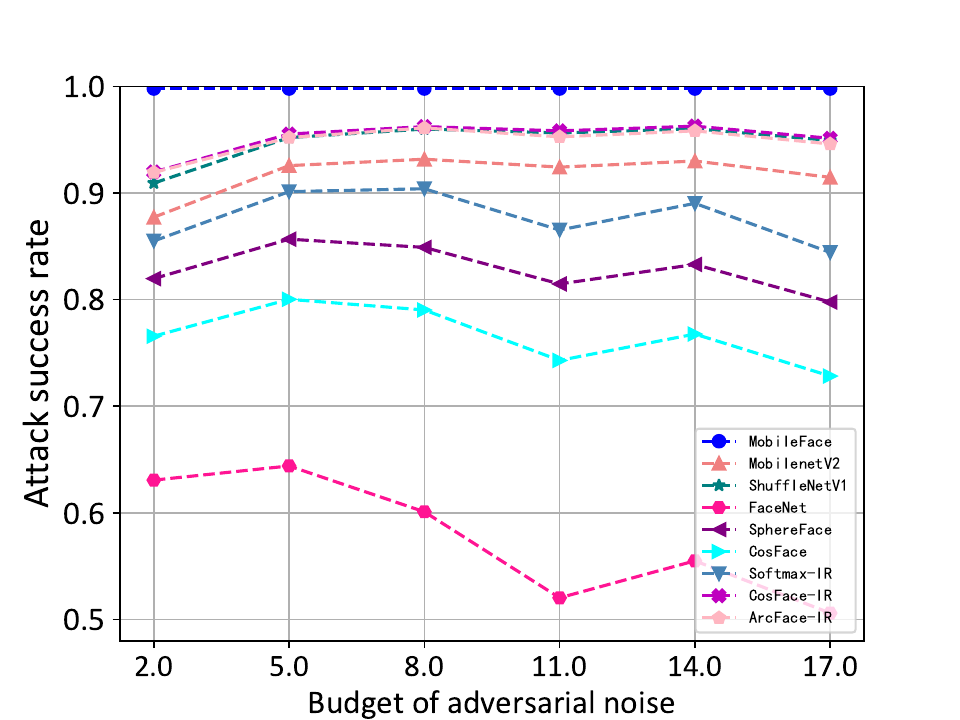}}~~\hfill~
    \subfigure[\textbf{Impersonation}]{
        \includegraphics[width=0.48\linewidth,height=0.35\linewidth]{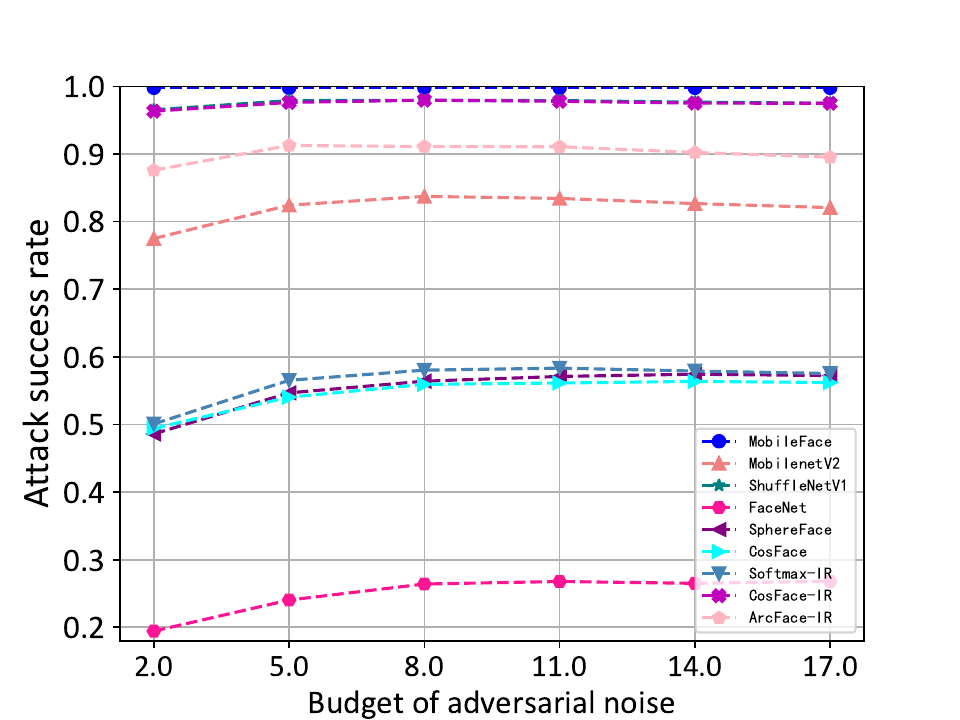}}~~\hfill~

    \caption{ \textbf{The influence of the budget of adversarial noise on attacking performance. ASRs against nine target models are exhibited for selecting a proper budget of adversarial noise. }}
    \label{budget_ablation}

\end{figure*}

\subsection{Ablation Study}
Considering the choice of layer to attack is crucial to realizing the utmost performance in feature-level attack, as verified in \cite{FIA,NAA}, it is necessary to explore attacking which layer (shallow, middle, or deep) is more likely to generate stronger adversarial examples against FR models that are characterized by handling the fine-grained task. Nine curves fluctuating with increasing layer index in Fig. \ref{layers_ablation} show that destroying feature maps approximating the middle layer is more likely to boost adversarial transferability. In addition to that, we also figured out how the budget of adversarial noise in our method is set to best benefit the attacking performance. The curves displayed in Fig. \ref{budget_ablation} suggest that a proper budget is important to achieve the utmost transferable attack, as a low budget of adversarial noise is inadequate to obscure the preferable features while a high budget is so excessive to pollute the face image that FR models can not accurately identify facial features and generate dubious transferred visual importance.

\section{Conclusion}
In this study, we consider the particularity of class-specific deep models for fined-grained vision tasks into adversarial attack for boosting the adversarial transferability. Taking the vision task of face verification as an example in this work, we observed that the decisive and auxiliary facial features are specific to each FR model, rather than following a similar biological mechanism of human visual system. This phenomenon indicates that the adversarial attack against a FR model inclines to destroy its decisive and auxiliary features, yet inevitably overfitting the adversarial examples to the model. To alleviate this problem, we propose the AAA, which aims to destroy the facial features that are critical for other FR models. We first try to transfer the attention of the source model using adversarial noise to imitate the attentions of other models on the clean face images, and then aggregate the obtained transferred attentions to imitate aggregation of attentions from various FR models for stronger adversarial examples. Extensive experimental results validate the superiority and robust effectiveness of our method over the baseline, and we further conduct the ablation study to probe the influence of two hyper-parameters in AAA on attacking performance, finding appropriate choices for them.



\renewcommand\refname{\zihao{5}\textbf{References}}


{
    \addtolength{\itemsep}{-2em}
    \zihao{5-}
    \bibliographystyle{IEEEtran}
    \bibliography{main}

\begin{thebibliography}{10}
\providecommand{\url}[1]{#1}
\csname url@samestyle\endcsname
\providecommand{\newblock}{\relax}
\providecommand{\bibinfo}[2]{#2}
\providecommand{\BIBentrySTDinterwordspacing}{\spaceskip=0pt\relax}
\providecommand{\BIBentryALTinterwordstretchfactor}{4}
\providecommand{\BIBentryALTinterwordspacing}{\spaceskip=\fontdimen2\font plus
\BIBentryALTinterwordstretchfactor\fontdimen3\font minus \fontdimen4\font\relax}
\providecommand{\BIBforeignlanguage}[2]{{%
\expandafter\ifx\csname l@#1\endcsname\relax
\typeout{** WARNING: IEEEtran.bst: No hyphenation pattern has been}%
\typeout{** loaded for the language `#1'. Using the pattern for}%
\typeout{** the default language instead.}%
\else
\language=\csname l@#1\endcsname
\fi
#2}}
\providecommand{\BIBdecl}{\relax}
\BIBdecl

\bibitem{quert-tar1}
A.~Ilyas, L.~Engstrom, A.~Athalye, and J.~Lin, ``Black-box adversarial attacks with limited queries and information,'' in \emph{International conference on machine learning}.\hskip 1em plus 0.5em minus 0.4em\relax PMLR, 2018, pp. 2137--2146.

\bibitem{quert-tar2}
Y.~Dong, S.~Cheng, T.~Pang, H.~Su, and J.~Zhu, ``Query-efficient black-box adversarial attacks guided by a transfer-based prior,'' \emph{IEEE Transactions on Pattern Analysis and Machine Intelligence}, vol.~44, no.~12, pp. 9536--9548, 2021.

\bibitem{quert-tar3}
Y.~Shi, Y.~Han, Q.~Hu, Y.~Yang, and Q.~Tian, ``Query-efficient black-box adversarial attack with customized iteration and sampling,'' \emph{IEEE Transactions on Pattern Analysis and Machine Intelligence}, vol.~45, no.~2, pp. 2226--2245, 2022.

\bibitem{DMA}
D.~Lin, Y.-G. Wang, W.~Tang, and X.~Kang, ``Boosting query efficiency of meta attack with dynamic fine-tuning,'' \emph{IEEE Signal Processing Letters}, vol.~29, pp. 2557--2561, 2022.

\bibitem{DML}
C.~Hu, H.-Q. Xu, and X.-J. Wu, ``Substitute meta-learning for black-box adversarial attack,'' \emph{IEEE Signal Processing Letters}, vol.~29, pp. 2472--2476, 2022.

\bibitem{DIM}
C.~Xie, Z.~Zhang, Y.~Zhou, S.~Bai, J.~Wang, Z.~Ren, and A.~L. Yuille, ``Improving transferability of adversarial examples with input diversity,'' in \emph{Proceedings of the IEEE/CVF conference on computer vision and pattern recognition}, 2019, pp. 2730--2739.

\bibitem{NI-SIM}
J.~Lin, C.~Song, K.~He, L.~Wang, and J.~E. Hopcroft, ``Nesterov accelerated gradient and scale invariance for adversarial attacks,'' \emph{arXiv preprint arXiv:1908.06281}, 2019.

\bibitem{FIA}
Z.~Wang, H.~Guo, Z.~Zhang, W.~Liu, Z.~Qin, and K.~Ren, ``Feature importance-aware transferable adversarial attacks,'' in \emph{Proceedings of the IEEE/CVF international conference on computer vision}, 2021, pp. 7639--7648.

\bibitem{NAA}
J.~Zhang, W.~Wu, J.-t. Huang, Y.~Huang, W.~Wang, Y.~Su, and M.~R. Lyu, ``Improving adversarial transferability via neuron attribution-based attacks,'' in \emph{Proceedings of the IEEE/CVF Conference on Computer Vision and Pattern Recognition}, 2022, pp. 14\,993--15\,002.

\bibitem{LGC}
X.~Yang, D.~Yang, Y.~Dong, H.~Su, W.~Yu, and J.~Zhu, ``Robfr: Benchmarking adversarial robustness on face recognition,'' \emph{arXiv preprint arXiv:2007.04118}, 2020.

\bibitem{face-photo}
G.~Davies, H.~Ellis, and J.~Shepherd, ``Cue saliency in faces as assessed by the ‘photofit’technique,'' \emph{Perception}, vol.~6, no.~3, pp. 263--269, 1977.

\bibitem{TIM}
Y.~Dong, T.~Pang, H.~Su, and J.~Zhu, ``Evading defenses to transferable adversarial examples by translation-invariant attacks,'' in \emph{Proceedings of the IEEE/CVF Conference on Computer Vision and Pattern Recognition}, 2019, pp. 4312--4321.

\bibitem{MIM}
Y.~Dong, F.~Liao, T.~Pang, H.~Su, J.~Zhu, X.~Hu, and J.~Li, ``Boosting adversarial attacks with momentum,'' in \emph{Proceedings of the IEEE conference on computer vision and pattern recognition}, 2018, pp. 9185--9193.

\bibitem{FGSM}
I.~J. Goodfellow, J.~Shlens, and C.~Szegedy, ``Explaining and harnessing adversarial examples,'' \emph{arXiv preprint arXiv:1412.6572}, 2014.

\bibitem{BIM}
A.~Kurakin, I.~J. Goodfellow, and S.~Bengio, ``Adversarial examples in the physical world,'' in \emph{Artificial intelligence safety and security}.\hskip 1em plus 0.5em minus 0.4em\relax Chapman and Hall/CRC, 2018, pp. 99--112.

\bibitem{PGN}
Z.~Ge, H.~Liu, W.~Xiaosen, F.~Shang, and Y.~Liu, ``Boosting adversarial transferability by achieving flat local maxima,'' \emph{Advances in Neural Information Processing Systems}, vol.~36, pp. 70\,141--70\,161, 2023.

\bibitem{STM}
Z.~Ge, F.~Shang, H.~Liu, Y.~Liu, L.~Wan, W.~Feng, and X.~Wang, ``Improving the transferability of adversarial examples with arbitrary style transfer,'' in \emph{Proceedings of the 31st ACM International Conference on Multimedia}, 2023, pp. 4440--4449.

\bibitem{qinghua1}
W.~Xue, X.~Xia, P.~Wan, P.~Zhong, and X.~Zheng, ``Adversarial attack on object detection via object feature-wise attention and perturbation extraction,'' \emph{Tsinghua Science and Technology}, 2024.

\bibitem{mid-layer-similarity}
M.~Salzmann \emph{et~al.}, ``Learning transferable adversarial perturbations,'' \emph{Advances in Neural Information Processing Systems}, vol.~34, pp. 13\,950--13\,962, 2021.

\bibitem{holistic}
V.~Bruce, P.~J. Hancock, and A.~M. Burton, ``Human face perception and identification,'' \emph{Face Recognition: From theory to applications}, pp. 51--72, 1998.

\bibitem{Resnet}
K.~He, X.~Zhang, S.~Ren, and J.~Sun, ``Identity mappings in deep residual networks,'' in \emph{Computer Vision--ECCV 2016: 14th European Conference, Amsterdam, The Netherlands, October 11--14, 2016, Proceedings, Part IV 14}.\hskip 1em plus 0.5em minus 0.4em\relax Springer, 2016, pp. 630--645.

\bibitem{mobilenetv2}
M.~Sandler, A.~Howard, M.~Zhu, A.~Zhmoginov, and L.-C. Chen, ``Mobilenetv2: Inverted residuals and linear bottlenecks,'' in \emph{Proceedings of the IEEE conference on computer vision and pattern recognition}, 2018, pp. 4510--4520.

\bibitem{LFW}
G.~B. Huang, M.~Mattar, T.~Berg, and E.~Learned-Miller, ``Labeled faces in the wild: A database forstudying face recognition in unconstrained environments,'' in \emph{Workshop on faces in'Real-Life'Images: detection, alignment, and recognition}, 2008.

\bibitem{FaceNet}
F.~Schroff, D.~Kalenichenko, and J.~Philbin, ``Facenet: A unified embedding for face recognition and clustering,'' in \emph{Proceedings of the IEEE conference on computer vision and pattern recognition}, 2015, pp. 815--823.

\bibitem{sphereface}
W.~Liu, Y.~Wen, Z.~Yu, M.~Li, B.~Raj, and L.~Song, ``Sphereface: Deep hypersphere embedding for face recognition,'' in \emph{Proceedings of the IEEE conference on computer vision and pattern recognition}, 2017, pp. 212--220.

\bibitem{cosface}
H.~Wang, Y.~Wang, Z.~Zhou, X.~Ji, D.~Gong, J.~Zhou, Z.~Li, and W.~Liu, ``Cosface: Large margin cosine loss for deep face recognition,'' in \emph{Proceedings of the IEEE conference on computer vision and pattern recognition}, 2018, pp. 5265--5274.

\bibitem{deng2019arcface}
J.~Deng, J.~Guo, N.~Xue, and S.~Zafeiriou, ``Arcface: Additive angular margin loss for deep face recognition,'' in \emph{Proceedings of the IEEE/CVF conference on computer vision and pattern recognition}, 2019, pp. 4690--4699.

\bibitem{MobileFace}
S.~Chen, Y.~Liu, X.~Gao, and Z.~Han, ``Mobilefacenets: Efficient cnns for accurate real-time face verification on mobile devices,'' in \emph{Biometric Recognition: 13th Chinese Conference, CCBR 2018, Urumqi, China, August 11-12, 2018, Proceedings 13}.\hskip 1em plus 0.5em minus 0.4em\relax Springer, 2018, pp. 428--438.

\bibitem{shufflenetv1}
X.~Zhang, X.~Zhou, M.~Lin, and J.~Sun, ``Shufflenet: An extremely efficient convolutional neural network for mobile devices,'' in \emph{Proceedings of the IEEE conference on computer vision and pattern recognition}, 2018, pp. 6848--6856.

\bibitem{smoothgrad}
D.~Smilkov, N.~Thorat, B.~Kim, F.~Vi{\'e}gas, and M.~Wattenberg, ``Smoothgrad: removing noise by adding noise,'' \emph{arXiv preprint arXiv:1706.03825}, 2017.

\bibitem{resnet52}
K.~He, X.~Zhang, S.~Ren, and J.~Sun, ``Deep residual learning for image recognition,'' in \emph{Proceedings of the IEEE conference on computer vision and pattern recognition}, 2016, pp. 770--778.

\end{thebibliography}
}

  \end{document}